\documentclass[conference]{IEEEtran}
\IEEEoverridecommandlockouts
\usepackage{cite}
\usepackage{amsmath,amssymb,amsfonts}
\usepackage{graphicx}
\usepackage{textcomp}
\usepackage{xcolor}
\usepackage{enumitem}
\usepackage{algorithm}

\usepackage[noend]{algpseudocode}

\usepackage{amsthm}
\usepackage{subcaption}
\newcommand{\ve}[1]{{\mbox{\boldmath${#1}$}}}
\theoremstyle{definition}

\usepackage{balance}

\def\BibTeX{{\rm B\kern-.05em{\sc i\kern-.025em b}\kern-.08em
    T\kern-.1667em\lower.7ex\hbox{E}\kern-.125emX}}
\begin{document}
\title{A Non-linear Function-on-Function Model for Regression with Time Series Data}
\author{
    \IEEEauthorblockN{Qiyao Wang\IEEEauthorrefmark{1}, Haiyan Wang\IEEEauthorrefmark{1}, Chetan Gupta\IEEEauthorrefmark{1},  Aniruddha Rajendra Rao\IEEEauthorrefmark{2}, Hamed Khorasgani\IEEEauthorrefmark{1}}
    \IEEEauthorblockA{\IEEEauthorrefmark{1}Industrial AI Lab, Hitachi America, Ltd. R$\&$D, Santa Clara, CA
    \\\{Qiyao.Wang, Haiyan.Wang, Chetan.Gupta, Hamed.Khorasgani\}@hal.hitachi.com}
   \IEEEauthorblockA{\IEEEauthorrefmark{2}Department of Statistics, Penn State University, University Park, PA
   \\\{arr30\}@psu.edu}
}

\maketitle
\footnotetext[1]{\copyright 2020 IEEE. Personal use of this material is permitted.  Permission from IEEE must be obtained for all other uses, in any current or future media, including reprinting/republishing this material for advertising or promotional purposes, creating new collective works, for resale or redistribution to servers or lists, or reuse of any copyrighted component of this work in other works.}

\begin{abstract}
In the last few decades, building regression models for non-scalar variables, including time series, text, image, and video, has attracted increasing interests of researchers from the data analytic community. In this paper, we focus on a multivariate time series regression problem. Specifically, we aim to learn mathematical mappings from multiple chronologically measured numerical variables within a certain time interval $\mathcal{S}$ to multiple numerical variables of interest over time interval $\mathcal{T}$. Prior arts, including the multivariate regression model, the Seq2Seq model, and the functional linear models, suffer from several limitations. The first two types of models can only handle regularly observed time series. Besides, the conventional multivariate regression models tend to be biased and inefficient, as they are incapable of encoding the temporal dependencies among observations from the same time series. The sequential learning models explicitly use the same set of parameters along time, which has negative impacts on accuracy. The function-on-function linear model in functional data analysis (a branch of statistics) is insufficient to capture complex correlations among the considered time series and suffer from underfitting easily. In this paper, we propose a general functional mapping that embraces the function-on-function linear model as a special case. We then propose a non-linear function-on-function model using the fully connected neural network to learn the mapping from data, which addresses the aforementioned concerns in the existing approaches. For the proposed model, we describe in detail the corresponding numerical implementation procedures. The effectiveness of the proposed model is demonstrated through the application to two real-world problems. 
\end{abstract}


\section{Introduction}\label{sec1}
In data analytics, one of the most important types of analysis is regression. The objective of regression is to mathematically estimate the relationship between one or more dependent variables (i.e., outcome variables being studied) and a set of independent variables (i.e., variables that have impacts on the outcome variables) \cite{kutner2005applied}. The learned mathematical mapping plays an effective role in not only sorting out which predictors/covariates, and how they interact with each other, to impact the dependent variables, but as well as predicting the outcomes for new samples. Due to its explanatory and predictive power, regression is an active area of research. Particularly, in the last few decades, building regression models for non-scalar variables, including time series, text, image, and video, has attracted increasing interests of researchers from the data analytic community. These modern regression models are beneficial to various domains in today's world where non-scalar types of data have become prevalent \cite{hochreiter1997long, fukushima1982neocognitron, chiou2016multivariate, xingjian2015convolutional}. 

In this paper, we consider the problem of regression with time series data that occur ubiquitously in many industrial and scientific fields. Specifically, we focus on building mathematical mappings from multiple chronologically measured numerical variables within a certain time interval $\mathcal{S}$ to multiple numerical variables of interest over time interval $\mathcal{T}$. It is noteworthy that depending on the data collecting mechanism, the time series can be either regular (i.e., the spacing of observation time is constant) or irregular (i.e., the spacing is not constant). Both types of time series are frequently encountered in real-world applications. Regular time series data is common in domains such as economics and meteorology. Examples include the daily temperature and the monthly interest rate. Irregularly spaced time series also naturally occur in many fields. For example, in the Internet of things, sensors often collect and transmit data only when the operational setting or the state of equipment changes, to reduce data storage and communication costs. In this paper, we consider the most general setting where the input and output time series could be either regular or irregular in the regression model. Furthermore, the data collection time could vary across variables and data instances. 

When data is irregular, a common practice is to preliminarily transform the time series into observations at a common and equally spaced time grid for all the subjects, so that the input and output data can be written as vectors consisting of temporally ordered scalar variables. Then the conventional multivariate regression models \cite{hair1998multivariate} or the sequence to sequence learning models \cite{hochreiter1997long, sutskever2014sequence} can be used to construct the mapping between the vectors of covariates and responses. A significant drawback of this approach is that the data manipulating step can introduce unquantifiable biases. Therefore, it is clear that regression models that directly use the unevenly spaced time series in their unaltered form would be useful. 

Besides, even for regular time series that don't necessitate pre-processing data into equally spaced time series, the existing approaches are known to suffer from several limitations. Despite their power in dealing with multiple variables observed at a single timestamp, the traditional multivariate regression models are inefficient in capturing temporal patterns and therefore, they are not good choices for solving regression problems with time series data. In particular, these models ignore the crucial fact that the covariates and responses consist of random observations of the same variables at different timestamps and solely rely on the regression models to account for the intricate temporal correlations. The sequential deep learning models are specifically designed to encode the sequential information in data with orders \cite{hochreiter1997long, jordan1990attractor, connor1994recurrent, dorffner1996neural}. These models have been widely deployed in difficult learning tasks in neural language processing \cite{hochreiter1997long, jordan1990attractor} and they have started to play an increasingly important role in time series data analysis in recent years \cite{connor1994recurrent, dorffner1996neural}. However, the sequential learning models are built upon iterating the same transformation on the hidden states (i.e., up-to-present memory) and the present covariates along time. This is a significant limitation, since the correlation between the responses and the covariates often varies over time. 

In the statistical field, function-on-function linear models (FFLM) are standard approaches for building regression models with time series covariates and responses \cite{faraway1997regression, chiou2016multivariate,FDA}. Under the central assumption that the time series are smooth realizations of underlying continuous stochastic processes, FFLM considers the entire time series as individual samples of the corresponding random processes and attempts to learn the unknown bivariate parameter function $\beta(s,t)$ that quantifies the correlation of covariates at any time $s\in \mathcal{S}$ with the response at any time $t\in \mathcal{T}$. This new perspective of modeling addresses the above-mentioned concerns in the previous approaches, since both regular and irregular time series can be analyzed as long as they contain a sufficient amount of information regarding the underlying continuous random process \cite{ramsay2006functional, yao2005functional,FDA, wang2017two}. More critically, unlike the sequential learning models, the flexible functional setting (i.e., $\beta(s,t)$) allows the correlation between covariates and responses to change within the considered time domains. However, these models are linear and therefore suffer from underfitting when the underlying mapping is complex.


In this paper, we also formulate the problem from the functional data analysis perspective. We innovatively identify a general mathematical mapping between the functional input and output data, based on which a non-linear model is proposed. The contributions of this paper are summarized as follows:

\begin{enumerate}[leftmargin=*]
    \item We propose a general functional mapping from multivariate temporal covariates to responses that embraces the function-on-function linear model as a special case. 
    \item We propose a new model to address the considered regression problem. For scenarios where the underlying process that generates the observed time series is smooth, the proposed model possesses several advantages, including its ability to handle versatile format of time series data, capture timely varying correlations among variables, and build complex mappings. 
    \item We describe in detail how to implement the proposed model from the beginning to the end and point to the existing packages that can be used in each step. 
    \item We demonstrate the effectiveness of the proposed approach through numerical experiments and apply it to solve two real-world challenges.  
\end{enumerate}

\section{Preliminaries}\label{sec2}
\subsection{Notations and Prior Art}\label{sec2.1}
The goal of the considered multivariate time series regression problem is to build a mapping from multiple time series covariates to several temporally measured responses, leveraging the temporal dependencies within and between the involved variables. 

Suppose that we have access to data from $N$ independent subjects. For each subject $i\in \{1,2,...,N\}$, $R$ covariates are continuously recorded within a compact time interval $\mathcal{S}\subseteq \mathbb{R}$. Note that subject and variable indexes are included in the following notations to reflect the fact that the measuring timestamps can vary across different variables and different subjects. In particular, the measuring timestamps of the $r$-th feature for subject $i$ are stored in a $M^{(i,r)}_s$-dimensional vector $\mathbf{S}^{(i,r)}=[S^{(i,r)}_{1},...,S^{(i,r)}_{j},...,S^{(i,r)}_{M^{(i,r)}_s}]^T$, with $M^{(i,r)}_s$ representing the number of observations in the time series and $S^{(i,r)}_{j} \in \mathcal{S}$ for $i=1,...,n; r=1,...,R; j=1,...,M^{(i,r)}_s$. The corresponding temporal observations are denoted as $\mathbf{X}^{(i,r)}=[X^{(i,r)}_{1},...,X^{(i,r)}_{j},...,X^{(i,r)}_{M^{(i,r)}_s}]^T$. Likewise, for a given subject $i$, there are $D$ responses being continuously measured within a compact time interval $\mathcal{T}\subseteq \mathbb{R}$. The $d$-th response is evaluated at $M^{(i,d)}_t$ timestamps. The measuring times and the observations are respectively represented by $\mathbf{T}^{(i,d)}=[T^{(i,d)}_{1},...,T^{(i,d)}_{j},...,T^{(i,d)}_{M^{(i,d)}_t}]^T$ and $\mathbf{Y}^{(i,d)}=[Y^{(i,d)}_{1},...,Y^{(i,d)}_{j},...,Y^{(i,d)}_{M^{(i,d)}_t}]^T$, for $d=1,...,D$.  In summary, the observed data is $\{\mathbf{X}^{(i,1)}, ..., \mathbf{X}^{(i,R)}, \mathbf{Y}^{(i,1)}, ..., \mathbf{Y}^{(i,D)}\}_{i=1}^N$. Intuitively, to effectively correlate the multivariate covariates over $\mathcal{S}$ to the responses over $\mathcal{T}$, it is required that, for any temporal covariate/response, there exist data from some subjects at timestamps across the period $\mathcal{S}$/$\mathcal{T}$, so that the overall temporal pattern can be estimated given data from the $N$ samples. Theoretical arguments that specify handleable irregularities for most of the data analytics models are provided in \cite{yao2005functional, paul2009consistency}. 

In the prior art, the time series need to be first transformed into regular time series evaluated at common time grids for all subjects. Let $M_s$ be the number of observations in the transformed covariates and $\mathbf{\tilde{X}}^{(i,r)}$ be the $M_s$-dimensional vector that stores the processed data for the $r$-th covariate of subject $i$. Similarly, let $M_t$ be the number of observations in the transformed responses and $\mathbf{\tilde{Y}}^{(i,d)}$ represents the $M_t$ observations from the $d$-th response of subject $i$. The conventional approaches then concatenate the $R$ temporal covariates and the $D$ time series-type responses, obtaining $\mathbf{\tilde{X}}^{(i)}=[\mathbf{\tilde{X}}^{{(i,1)}^T},...,\mathbf{\tilde{X}}^{{(i,R)}^T}]^T$  and $\mathbf{\tilde{Y}}^{(i)}=[\mathbf{\tilde{Y}}^{{(i,1)}^T},...,\mathbf{\tilde{Y}}^{{(i,D)}^T}]^T$. Given samples $\{\mathbf{\tilde{X}}^{(i)}, \mathbf{\tilde{Y}}^{(i)}\}_{i=1}^N$, the multivariate regression or sequential learning models are then utilized to learn the mapping
\begin{equation} \label{formulation1}
\mathbf{\tilde{Y}}^{(i)} = F(\mathbf{\tilde{X}}^{(i)}). 
\end{equation}

The disadvantages of the above approaches are twofold. On one hand, immensurable biases may be introduced when conducting data pre-processing so that the learned mapping tends to deviate from the ground truth. On the other hand, these widely used models have their own limitations in solving regression problems with time series inputs and outputs. The multivariate regression models are incapable of encoding the temporal dependencies among $\mathbf{\tilde{X}}^{(i,r)}$ and $\mathbf{\tilde{Y}}^{(i,d)}$, $r=1,...,R$ and $d=1,...,D$. Accordingly, the sequential learning models explicitly apply the same mathematical operations (i.e., use the same set of parameters) on the $R$-dimensional inputs and the hidden states to obtain the $D$-dimensional outputs for all the timestamps.

\subsection{Functional Data Analysis and a New Formulation}\label{sec2.2}
In this section, we describe an alternative problem formulation from the functional data analysis (FDA) point of view. Functional data analysis refers to the analysis of data samples consisting of dynamically varying data over a continuum. It is a key methodology for the analysis of data that can be viewed as realizations of random functions or surfaces, such as time series, image, and tracking data (e.g., handwriting and driving path) \cite{ramsay2006functional}. When modeling time series data, FDA methods uniquely deal with the continuous underlying curves $X^{(i,r)}(s), s \in  \mathcal{S}$ that generate the observed discrete time series $\mathbf{X}^{(i,r)}$. The input and output data for functional regression models are $\{X^{(i,1)}(s), ..., X^{(i,R)}(s), s\in \mathcal{S}; Y^{(i,1)}(t), ..., Y^{(i,D)}(t), t\in \mathcal{T}\}_{i=1}^N$. Denote the sequential covariates and responses as vectors of random functions, i.e., $\mathbf{X}^{(i)}(s)=[X^{(i,1)}(s), ..., X^{(i,R)}(s)]^T$ and $\mathbf{Y}^{(i)}(t)=[Y^{(i,1)}(t), ..., Y^{(i,D)}(t)]^T$. Functional regression models aim to learn the mapping 
\begin{equation} \label{formulation2}
\mathbf{Y}^{(i)}(t) = F(\mathbf{X}^{(i)}(s)). 
\end{equation}
For instance, function-on-function linear models \cite{faraway1997regression, chiou2016multivariate,FDA} focus on learning the bivariate parameter functions in a $D \times R$ matrix $\ve{\beta}(s,t)=[\beta_{r,d}(s,t)]^T_{r=1,...,R; d=1,...,D}$
\begin{equation} \label{formulation3}
\mathbf{Y}^{(i)}(t) = \ve{\mu}(t) + \int_{s} \ve{\beta}(s,t) \mathbf{X}^{(i)}(s)ds,
\end{equation}
where $\ve{\mu}(t)=[\mu_1(t),...,\mu_D(t)]^T$ consists of the mean function for the $D$ responses over $\mathcal{T}$.

In FDA, although smoothness of underlying random functions $\{\mathbf{X}^{(i)}(s), s\in \mathcal{S}; \mathbf{Y}^{(i)}(t), t\in \mathcal{T}\}_{i=1}^N$, such as existence of continuous second derivatives, is often imposed for regularization, FDA techniques often accommodate moderate random errors in the actual discrete observations \cite{yao2005functional}. Therefore, they are applicable in analyzing a wide range of time series data. Unlike the conventional models in Section \ref{sec2.1}, functional models can directly analyze the raw time series, which greatly enhances the flexibility in applications. Furthermore, in contrast to the sequential learning models that keep the parameters the same over time, functional models abandon this restrictive assumption and explicitly allow the covariate effects to change along $\mathcal{S}$ for different timestamps in the response time interval $\mathcal{T}$. 

Most studies on regression models with both functional covariates and functional responses have focused on linear models \cite{faraway1997regression, chiou2014multivariate}. However, the linear structures are inadequate and make the models suffer from underfitting easily \cite{rossi2002functional, wang2019multilayer, wang2019remaining, rao2020spatio}. In the next section, we propose a non-linear function-on-function regression model leveraging the power of fully connected Neural Networks.

\section{Proposed Non-linear Function-on-function Regression Model}\label{sec3}
\subsection{Multivariate Functional principal Component Analysis}\label{sec3.1}
In the section, we briefly summarize the multivariate functional principal component analysis (multivariate FPCA), an useful dimension reduction tool that frequently serves as a key component in many functional models \cite{happ2018multivariate, chiou2014multivariate}. This section focuses on describing the theory. The specific estimation procedures using the actual observations are included in Section \ref{sec3.3}. 

The basic objects in multivariate FPCA is a set of real-valued random functions on a common compact interval, such as the multivariate functional covariates and responses introduced in Section \ref{sec2.2}. Let's take the $D$-dimensional functional responses $\mathbf{Y}^{(i)}(t)=[Y^{(i,1)}(t), ..., Y^{(i,D)}(t)]^T$ as an example. Each element in $\mathbf{Y}^{(i)}(t)$ is typically assumed to follow a stochastic process with unknown mean function $\mu_Y^{(d)}(t)$ and covariance function $G_Y^{(dd)}(t, t^\prime)$, for $d=1,...,D$. Also, the variables are cross-correlated, with the covariance function between the $d$-th and the $d^\prime$-th functional variable being $G_Y^{(dd^\prime)}(t, t^\prime)$, for $d,d^\prime = 1,....,D$ and $t, t^\prime \in \mathcal{T}$. 

To take the possibly uneven extent of variations among the $D$ random processes into account, we follow the proposal in \cite{chiou2014multivariate} to normalize data through a point-wise Z-score standarization, i.e., $Y_{z}^{(i,d)}(t)=v_Y^{(d)}(t)^{-1/2}(Y^{(i,d)}(t)-\mu_Y^{(d)}(t))$, with $v_Y^{(d)}(t)=G_Y^{(dd)}(t, t)$ being the variance among observations at time $t$. Let's denote the normalized random functions as  $\mathbf{Y}_z^{(i)}(t)=[Y_z^{(i,1)}(t), ..., Y_z^{(i,D)}(t)]^T$, whose mean is a $D$-dimensional function, which takes a value $0$ over $\mathcal{T}$ and the matrix of covariance functions is $\mathbf{G}_{Y_z}(t, t^\prime)=[G_{Y_z}^{(dd^\prime)}(t, t^\prime)]_{d, d^\prime =1,...,D}$, $t, t^\prime \in \mathcal{T}$. 

Under certain regularity requirements on $\mathbf{G}_{Y_z}(t, t^\prime)$, it has been shown that there exists an orthonormal basis of eigenfunctions $\ve{\phi}_p(t)=[\phi_p^{(1)}(t),...,\phi_p^{(D)}(t)]^T$ such that 
\begin{equation} \label{mfpca1}
\int \mathbf{G}_{Y_z}(t, t^\prime) \ve{\phi}_p(t^\prime)d t^\prime = \lambda_p \ve{\phi}_p(t),  \text{with } \lim_{p\rightarrow \infty}\lambda_p = 0,
\end{equation}
where $\lambda_p \in \mathbb{R}$ is the eigenvalue corresponding to the $D$-dimensional eigenfunction $\ve{\phi}_p(t)$. Similar to the conventional PCA, $\lambda_p$ quantifies the amount of variability in $\mathbf{Y}_z^{(i)}(t)$ being captured by $\ve{\phi}_p(t)$. The output of $\int \mathbf{G}_{Y_z}(t, t^\prime) \ve{\phi}_p(t^\prime)d t^\prime$ is a $D$-dimensional function over $\mathcal{T}$, with the $d$-th element $(\int \mathbf{G}_{Y_z}(t, t^\prime) \ve{\phi}_p(t^\prime)d t^\prime)^{(d)}$ being
\begin{equation} \label{mfpca2}
\left(\int \mathbf{G}_{Y_z}(t, t^\prime) \ve{\phi}_p(t^\prime)d t^\prime\right)^{(d)} = \sum_{d^\prime=1}^D \int G_{Y_z}^{(dd^\prime)}(t, t^\prime)\phi_p^{(d^\prime)}(t^\prime)d t^\prime.
\end{equation}

Given Eq \eqref{mfpca1}, it has been shown that the multivariate random function $\mathbf{Y}_z^{(i)}(t)$ can be represented as a linear combination of the multivariate eigenfunctions with real-valued random coefficients $\xi_p^{(i)} = \sum_{d=1}^D \int \phi_p^{(d)}(t) Y_z^{(i,d)}(t) dt$, $p=1,...,P$. Mathematically, it can be seen as 
\begin{equation} \label{mfpca3}
\mathbf{Y}_z^{(i)}(t) = \sum_{p=1}^\infty \xi_p^{(i)} \ve{\phi}_p(t).
\end{equation}
Due to the smoothness of each functions in $\mathbf{Y}_z^{(i)}(t)$, it has been proved that the eigenvalues decay to 0 at a fast rate such that the information in  $\mathbf{Y}_z^{(i)}(t)$ is well captured by a finite number of the random coefficients. That is Eq \eqref{mfpca3} becomes 
\begin{equation} \label{mfpca4}
\mathbf{Y}_z^{(i)}(t) \approx \sum_{p=1}^P \xi_p^{(i)} \ve{\phi}_p(t).
\end{equation}
The number of truncation $P$ is typically determined by methods such as fraction of variance explained, cross-validation, or some AIC and BIC-based approaches \cite{yao2005functional}.

In the next section, we describe in detail the proposed non-linear function-on-function regression model based on multivariate FPCA, followed by discussions on numerical implementations in Section \ref{sec3.3}.

\subsection{Proposed Model}\label{sec3.2}
Following the convention in the literature \cite{chiou2016multivariate, chiou2014multivariate}, let's first use the point-wise Z-score normalization to make the magnitude of variation comparable among the covariates and responses. Let's denote the standardized data as $\mathbf{X}_z^{(i)}(s)=[X_z^{(i,1)}(s), ..., X_z^{(i,R)}(s)]^T$ and $\mathbf{Y}_z^{(i)}(t)=[Y_z^{(i,1)}(t), ..., Y_z^{(i,D)}(t)]^T$. To learn the mapping between $\mathbf{X}_z^{(i)}(s)$ and $\mathbf{Y}_z^{(i)}(t)$, some existing models assume certain functional forms and then estimate the parameter functions associated with the functional covariates. For instance, the linear model being considered in literature is 
\begin{equation} \label{formulation4}
\mathbf{Y}_z^{(i)}(t) = \int_{s} \ve{\beta}(s,t) \mathbf{X}_z^{(i)}(s)ds.
\end{equation}

 Eq \eqref{mfpca4} implies that $\mathbf{Y}_z^{(i)}(t)$ can be effectively approximated by a linear combination of the eigenfunctions $\ve{\phi}_p(t)$, with the major modes of variations among $\mathbf{Y}_z^{(i)}(t)$ captured. Under the regression setting, to learn the impact of different values of $\mathbf{X}_z^{(i)}(s)$ on the variation of $\mathbf{Y}_z^{(i)}(t)$, we propose to set the coefficients $\xi_p^{(i)}$ in Eq \eqref{mfpca4} as a function of $\mathbf{X}_z^{(i)}(s)$. Mathematically, let $\ve{\Phi}$ be a $D\times P$ matrix, with the $(d,p)$-th element being $\phi_p^{(d)}(t)$. Let $\mathbf{f}(\cdot)$ be a mapping from $R$-dimensional $L_2(\mathcal{T})$ to $\mathbb{R}^P$. Then we define a general mapping from $\mathbf{X}_z^{(i)}(s)$ to $\mathbf{Y}_z^{(i)}(t)$ as 
\begin{equation} \label{formulation_new1}
\mathbf{Y}_z^{(i)}(t) = \ve{\Phi} \mathbf{f}(\mathbf{X}_z^{(i)}(s)).
\end{equation}

The goal of the regression problem is then to learn the mapping $\mathbf{f}(\cdot)$ in Eq \eqref{formulation_new1}. It is worth noting that the function-on-function linear model in Eq \eqref{formulation4} is a special case of the general mapping in Eq \eqref{formulation_new1}. The linear models set the unknown parameter function in Eq \eqref{formulation4} as $\ve{\beta}(s,t)= \ve{\Phi} \mathbf{B} \ve{\Psi}^T$, where $\ve{\Psi}$ is a $R\times L$ matrix that stores the first $L$ multivariate eigenfunctions of $\mathbf{X}_z^{(i)}(s)$ and $\mathbf{B}$ is a $P\times L$ matrix holds real-valued unknown parameters that defines the parameter function $\ve{\beta}(s,t)$ \cite{chiou2016multivariate, ramsay2006functional}. This is equivalent to say that these models consider $\mathbf{f}(\cdot)$ in Eq \eqref{formulation_new1} as $\mathbf{B} \ve{\Psi}^T\mathbf{X_z^{(i)}}(s)$. 

\begin{figure}[htbp]
	\centering
	\includegraphics[width=90mm]{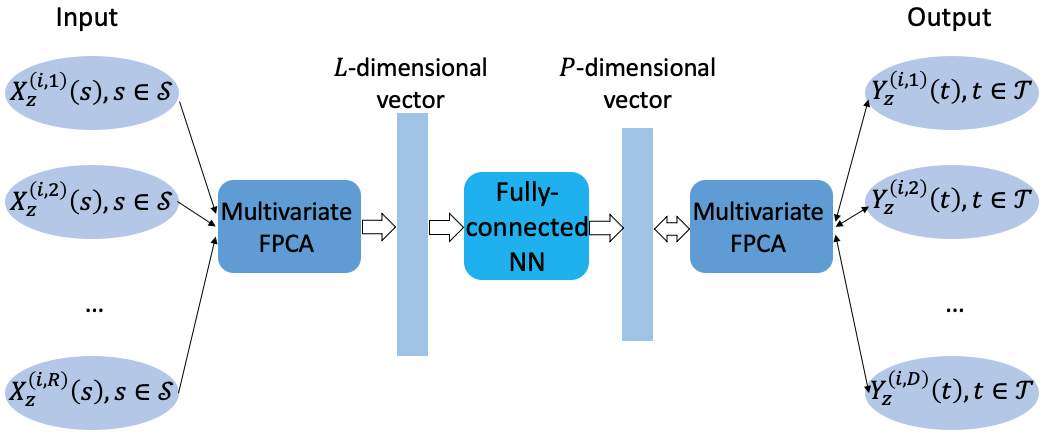} 
	\caption{A graphical representation of the proposed model. }\label{Fig:arch}
\end{figure}

To capture non-linear relationships, in this paper, we propose to replace the linear structure by a fully connected Neural Network with $W$ layers and the number of neurons being $O_w$, for $w=1,...,W$. Specifically, the model is
\begin{equation} \label{formulation_new2}
\mathbf{Y}_z^{(i)}(t) = \ve{\Phi} \textbf{NN}_{\{O_w\}_{w=1}^W}(\ve{\Psi}^T\mathbf{X}_z^{(i)}(s)).
\end{equation}
Leveraging the power of Neural Networks, the proposed mapping in Eq \eqref{formulation_new2} is able to learn an intricate mapping from the multivariate covariates $\mathbf{X}_z^{(i)}(s)$ to the response $\mathbf{Y}_z^{(i)}(t)$. This extension significantly enhances the applicability of functional regression models. Note that $W, O_1,...,O_W$ are hyperparameters that can be carefully tuned to obtain the best model performance. The graphical representation of the proposed mode is provided in Fig. \ref{Fig:arch}.

\subsection{Numerical Implementation}\label{sec3.3}
In this section, we consider the numerical implementation involved in the training and application phase of the proposed model. 

\subsubsection{Training phase}\label{sec3.3.1}
In this subsection, we present in details how the model in Eq \eqref{formulation_new2} can be trained using the finitely observed data $\{\mathbf{X}^{(i,1)}, ..., \mathbf{X}^{(i,R)}, \mathbf{Y}^{(i,1)}, ..., \mathbf{Y}^{(i,D)}\}_{i=1}^N$. 
\begin{itemize}[leftmargin=*] 

\item \textit{Data standardization:}  We begin by summarizing the estimation procedure for the mean and covariance functions for each of the considered variables. Let's use the $d$-th response as an example. Note that the same procedure will be iterated on the other variables. Given data $\{ \mathbf{Y}^{(i,d)}\}_{i=1}^N$ and the corresponding observation time $\{ \mathbf{T}^{(i,d)}\}_{i=1}^N$, for any $t \in \mathcal{T}$, let's define a smoothing kernel $K_{Y,1}^{(d)}(\cdot)$ with a tunable bandwidth  $h_{\mu_Y^{(d)}}$ and minimize
\begin{equation}
\resizebox{1.0\hsize}{!}{
$\sum_{i=1}^{N}\sum_{j=1}^{M_t^{(i,d)}}K_{Y,1}^{(d)}\left(\frac{T^{(i,d)}_{j}-t}{h_{\mu_Y^{(d)}}}\right)\left[Y^{(i,d)}_j-\beta^{(d)}_{Y,0}-\beta^{(d)}_{Y,1}(t-T^{(i,d)}_{j})\right]^2$}
\end{equation}
with respect to $\beta^{(d)}_{Y,0}$ and $\beta^{(d)}_{Y,1}$, leading  $\hat{\mu}_Y^{(d)}(t)=\hat{\beta}^{(d)}_{Y,0}(t)$. For $i=1,...,N$, let $U_Y^{(i,d)}(T^{(i,d)}_{j_1}, T^{(i,d)}_{j_2})=(Y^{(i,d)}_{j_1} - \hat{\mu}_Y^{(d)}(T^{(i,d)}_{j_1}))(Y^{(i,d)}_{j_2} - \hat{\mu}_Y^{(d)}(T^{(i,d)}_{j_2}))$. Also, let's define a smoothing kernel $K_{Y,2}^{(d)}(\cdot, \cdot)$ with a bandwidth $h_{G_Y^{(d)}}$. For any given $t, t^\prime \in \mathcal{T}$, minimize 
\begin{multline*}
\resizebox{0.85\hsize}{!}{$\sum_{i=1}^{N}\sum_{1\leq j_1 \neq j_2 \leq M_t^{(i,d)}}K_{Y,2}^{(d)}\left(\frac{T^{(i,d)}_{j_1}-t}{h_{G_Y^{(d)}}}, \frac{T^{(i,d)}_{j_2}-t^\prime}{h_{G_Y^{(d)}}}\right)\times$}\\
\resizebox{1.0\hsize}{!}{$\left[U_Y^{(i,d)}(T^{(i,d)}_{j_1}, T^{(i,d)}_{j_2})-\gamma^{(d)}_{Y,0}-\gamma^{(d)}_{Y,1}(t-T^{(i,d)}_{j_1}) - \gamma^{(d)}_{Y,2}(t^\prime-T^{(i,d)}_{j_2})\right]^2$}
\end{multline*}
with respect to $\gamma^{(d)}_{Y,0}$, $\gamma^{(d)}_{Y,1}$ and $\gamma^{(d)}_{Y,2}$, leading  $\hat{G}_Y^{(d)}(t, t^\prime)=\hat{\gamma}^{(d)}_{Y,0}(t, t^\prime)$ and accordingly, $\hat{v}_Y^{(d)}(t)=\hat{\gamma}^{(d)}_{Y,0}(t, t)$, for $t\in\mathcal{T}$. The estimation procedure can be implemented by R packages including `fdapace' \cite{fdapaceR} and `fpca'  \cite{fpcaR}. Given the estimated mean and covariance function, we obtain the standardized data $\{ \mathbf{Y}_{z}^{(i,d)}\}_{i=1}^N$, with the $j$-th element of $\mathbf{Y}_{z}^{(i,d)}$ being 
\begin{equation}
Y_{z,j}^{(i,d)}=\hat{v}_Y^{(d)}\left(T^{(i,d)}_{j}\right)^{-1/2}\left(Y^{(i,d)}_j-\hat{\mu}_Y^{(d)}\left(T^{(i,d)}_{j}\right)\right).
\end{equation}
After conducting the above calculation for all the variables, we obtain the standardized data $\{\mathbf{X}_{z}^{(i,1)}, ..., \mathbf{X}_{z}^{(i,R)}, \mathbf{Y}_{z}^{(i,1)}, ..., \mathbf{Y}_{z}^{(i,D)}\}_{i=1}^N$, with observation times being $\{\mathbf{S}^{(i,1)}, ..., \mathbf{S}^{(i,R)}, \mathbf{T}^{(i,1)}, ..., \mathbf{T}^{(i,D)}\}_{i=1}^N$ as defined in Section \ref{sec2.1}.

\item \textit{Multivariate FPCA}: The objective is to estimate the multivariate functional principal components $\ve{\Phi}$ and $\ve{\Psi}$ in the proposed model in Eq \eqref{formulation_new2}. The core theoretical result in \cite{happ2018multivariate} implies that the multivariate principal components $\ve{\Phi}$ and $\ve{\Psi}$ can be represented by the less involved univariate functional principal components. In particular, for $d=1,...,D$, we start with using data $\{ \mathbf{Y}_{z}^{(i,d)}\}_{i=1}^N$ to obtain the estimated univariate eigenfunction $\{\hat{\phi}_p^*{}^{(d)}(t)\}$ through
the restricted maximum likelihood method \cite{peng2009geometric} or the local linear smoothing based approach \cite{yao2005functional}, both of which are included in R packages `fdapace' and `fpca'. Next, we estimate $\int Y_z^{(i,d)}(t) \phi_p^*{}^{(d)}(t)dt$ based on $\mathbf{Y}_{z}^{(i,d)}$ and $\hat{\phi}_p^*{}^{(d)}(t)$ through numerical integration \cite{davis2007methods}. Let's denote the achieved value as $\hat{\xi}_p^*{}^{(i,d)}$, for $i=1,...,N; d=1,...,D; p=1,...,P_d$. Let $P_+=\sum_{d=1}^D P_d$ and $\ve{\Xi}$ is a $P_+ \times P_+$ consisting of blocks $\ve{\Xi^{(dd^\prime)}} \in \mathbb{R} ^{P_d\times P_{d^\prime}}$ with the $(p, p^\prime)$-th entry being
\begin{equation}
\begin{split}
\Xi_{pp^\prime}^{(dd^\prime)} & = \text{Cov}(\hat{\xi}_p^*{}^{(i,d)}, \hat{\xi}_{p^\prime}^*{}^{(i,d^\prime)})\\
& = \frac{1}{N-1}\sum_{i=1}^N (\hat{\xi}_p^*{}^{(i,d)} - \delta_{p}^*{}^{(d)})(\hat{\xi}_{p^\prime}^*{}^{(i,d^\prime)} - \delta_{p^\prime}^*{}^{(d^\prime)}),
\end{split}
\end{equation}
where $\delta_{p}^*{}^{(d)}$=$\frac{1}{N}\sum_{i=1}^N\hat{\xi}_p^*{}^{(i,d)}$ and $\delta_{p^\prime}^*{}^{(d^\prime)}=\frac{1}{N}\sum_{i=1}^N\hat{\xi}_{p^\prime}^*{}^{(i,d^\prime)} $ are the corresponding sample means. Let's conduct eigen decomposition on matrix $\ve{\Xi}$ and denote the $p$-th eigenvector as $\mathbf{c}_p$. Note that $\mathbf{c}_p$ can be considered as a vector consisting of $D$ blocks, with the $d$-th block being denoted as $[\mathbf{c}_p]^{(d)} \in \mathbb{R}^{P_d}$. According to the proposition in \cite{happ2018multivariate}, we can estimate the $(d, p)$-th element of $\ve{\Phi}$ (i.e., the matrix of the multivariate functional principal components) in Section \ref{sec3.2} by
\begin{equation}
\hat{\phi}_p^{(d)}(t) = \sum_{m=1}^{P_d}[\mathbf{c}_p]_m^{(d)} \hat{\phi}_m^*{}^{(d)}(t).
\end{equation}
Likewise, we can estimate the multivariate functional principal components of the $R$-dimensional covariates. Let's denote the achieved estimate as $\ve{\hat{\Phi}}$ and $\ve{\hat{\Psi}}$. 

\item \textit{Train the proposed model}: Given the orthnormality of eigenfunctions, Eq \eqref{formulation_new2} is equivalent to 
\begin{equation} \label{formulation_new22}
\ve{\Phi}^T \mathbf{Y}_z^{(i)}(t) =  \textbf{NN}_{\{O_w\}_{w=1}^W}(\ve{\Psi}^T\mathbf{X}_z^{(i)}(s)). 
\end{equation}
Given this observation, to train the proposed model, we first use numerical integration to estimate the $P$-dimensional output vector $\ve{\Phi}^T \mathbf{Y}_z^{(i)}(t)$ from $\ve{\hat{\Phi}}$ and $\{\mathbf{Y}_z^{(i,1)}, ..., \mathbf{Y}_z^{(i,D)}\}$. Similarly, we estimate the $L$-dimensional input vector $\ve{\Psi}^T \mathbf{X}_z^{(i)}(s)$. Next, a fully connected neural network is trained based on the estimated scalar projections of $\ve{\Phi}^T \mathbf{Y}_z^{(i)}(t)$ and $\ve{\Psi}^T \mathbf{X}_z^{(i)}(s)$. 
\end{itemize}

\subsubsection{Application phase}\label{sec3.3.2}
In this subsection, we describe the application of the trained model to a new subject $\{\mathbf{X}^{(new,1)}, ..., \mathbf{X}^{(new,R)}\}$, with $\mathbf{X}^{(new,r)}$ consisting of discretized observations of the $r$-th feature within time interval $\mathcal{S}$. The observation times are denoted as $\mathbf{S}^{(new,r)}$, for $r=1,...,R$.

\begin{itemize}[leftmargin=*] 
\item \textit{Data standardization:} To deploy the learned model on the new data, we first conduct data standardization for each of the covariates. For $r=1,....,R$, let the estimated mean and variance function from the input data in training be $\hat{\mu}_X^{(r)}(s)$ and $\hat{v}_X^{(r)}(s)$, the $j$-th element of the standardized data $\mathbf{X}_{z}^{(new,r)}$ is 
\begin{equation}\label{norm_test_data}
\resizebox{0.95\hsize}{!}{$X_{z,j}^{(new,r)}=\hat{v}_X^{(r)}\left(S^{(new,r)}_{j}\right)^{-1/2}\left(X^{(new,r)}_j-\hat{\mu}_X^{(r)}\left(S^{(new,r)}_{j}\right)\right)$}.
\end{equation}
The normalized data is $\{\mathbf{X}_{z}^{(new,1)}, ..., \mathbf{X}_{z}^{(new,R)}\}$.

\item \textit{Make predictions:} Given the normalized data in  Eq \eqref {norm_test_data}, $\ve{\hat{\Psi}}$, and the multivariate FPC learned in the training phase, we can compute estimations for the $L$-dimensional vector  $\ve{\Psi}^T \mathbf{X}_z^{(new)}(s)$, where $\mathbf{X}_z^{(new)}(s)$ are the underlying continuous functions that render the discrete observations $\{\mathbf{X}_{z}^{(new,1)}, ..., \mathbf{X}_{z}^{(new,R)}\}$. Let's denote the achieved estimation as $\ve{\eta}^{(new)}$, then we have 
\begin{equation} \label{prediction}
\mathbf{\hat{Y}}_z^{(new)}(t) =  \ve{\hat{\Phi}}\hat{\textbf{NN}}_{\{O_w\}_{w=1}^W}(\ve{\eta}^{(new)}).
\end{equation}
Note that $\mathbf{\hat{Y}}_z^{(new)}(t)$ is a $D$-dimensional vector of functions, with the $d$-th element corresponding to the prediction of the $d$-th response variable given $\{\mathbf{X}_{z}^{(new,1)}, ..., \mathbf{X}_{z}^{(new,R)}\}$.

\item \textit{Convert the prediction to its original scale:} In this section, for each prediction in $\mathbf{\hat{Y}}_z^{(new)}(t)$, we revert the calculation in the point-wise Z-score standardization to scale it back to its original extent. Mathematically, the final estimate $\hat{Y}^{(new,d)}(t)$ is 
\begin{equation} \label{prediction2}
\hat{Y}^{(new,d)}(t) =  \hat{Y}^{(new,d)}(t) \hat{v}_Y^{(d)}(t)^{1/2} + \hat{\mu}_Y^{(d)}(t),
\end{equation}
where $\hat{\mu}_Y^{(d)}(t)$ and $\hat{v}_Y^{(d)}(t)$  are the mean and variance function estimates from the training data. 

\end{itemize}

\section{Numerical Experiments}\label{sec4}
In this section, we apply the proposed model to solve two real-world challenges. The first problem is about understanding the association between electricity demand and temperature, which plays a key role in electricity supply management. The second problem attempts to perform short-term traffic prediction for facilitating driving decision making and improving the overall transportation efficiency. We compare the performance of the proposed non-linear functional model with several state-of-the-art approaches, including the multivariate regression model (specifically, the multivariate linear regression), the widely used Seq2Seq model with LSTM-based encoder and decoder \cite{sutskever2014sequence}, and the function-on-function linear regression model \cite{chiou2016multivariate}. As shown by the experimental results, the proposed model can serve as an effective solution for regressions with multivariate temporal covariates and responses. For the considered problems, the proposed model outperforms the above-mentioned alternative methods.   

\subsection{An Application to Electricity Demand Analysis}\label{sec4.1}
In the energy field, due to the high costs associated with the storage of electricity, it is of great significance to understand the variability of electricity demand within a region over time so that the authorities and suppliers can make informative operational decisions. Temperature is one of the most essential factors that contribute to the variability in electricity demand, as the weather condition affects the usage of heating and cooling appliances. In the literature, several efforts have been made to investigate the association between temperature and electricity demand \cite{liu2020multivariate, magnano2007mathematical}. In this experiment,  following the formulation in \cite{liu2020multivariate}, we apply the proposed model to build a mathematical mapping from the daily temperature trajectory for the 7 days of a week to the daily electricity demand trajectory  for the 7 days of the same week. 

\begin{figure*}[htbp]
	\centering
	\begin{subfigure}[t]{7.05in}
		\centering
        \caption{7-dimensional covariates in the sample.}\label{temp}
		\includegraphics[width=18.25cm]{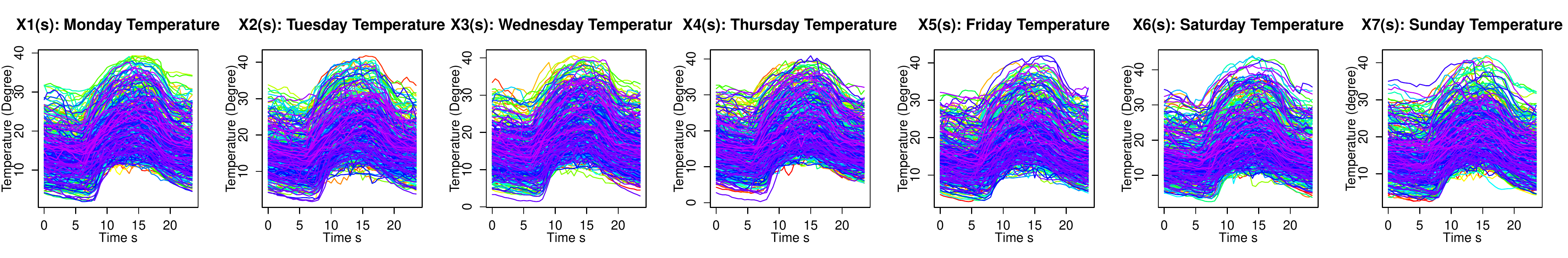}		
	\end{subfigure}
    \par
	\begin{subfigure}[t]{7.05in}
		\centering
        \caption{7-dimensional responses in the sample.}\label{demand}
		\includegraphics[width=18.25cm]{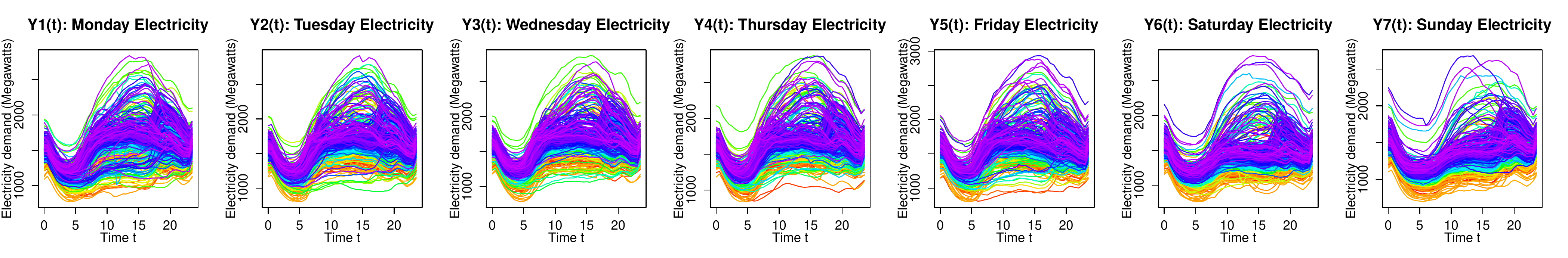}		
	\end{subfigure}
	\vspace{-0.05in}
	\caption{Observed within-a-day trajectories of temperature and electricity demand for the 508 weeks in the experiment.}\label{dataset1}
\end{figure*}

We use the temperature and electricity demand records of Adelaide, a city in the state of South Australia, between 7/6/1997 and 3/31/2007. For a given day, we have access to half-hourly data for both temperature and electricity demand, i.e. the observed covariates and responses are regular time series of length 48 within the 24 hours time period. There are 508 weeks within the considered period, i.e., sample size $N=508$. In Fig. \ref{dataset1}, we plot the multivariate temporal covariates and responses for all 508 samples. 


We randomly select 400 samples to build the regression model and evaluate the model performance on the remaining 108 weeks of data. Let the number of subjects in the testing set be $N_{\text{test}}$. The ground truth and estimation for the half-hourly electricity demand be $\mathbf{Y}^{(i,d)}=[Y^{(i,d)}_1,...,Y^{(i,d)}_j,...,Y^{(i,d)}_{48}]$ and $\mathbf{\hat{Y}}^{(i,d)}=[\hat{Y}^{(i,d)}_1,...,\hat{Y}^{(i,d)}_j,...,\hat{Y}^{(i,d)}_{48}]$ for $i=1,...,N_{\text{test}}; d=1,...,48$. For a given response, we quantify the accuracy by the root mean squared error (RMSE)
\begin{equation} \label{metric1}
\resizebox{1.0\hsize}{!}{$\text{RMSE}\left(\mathbf{Y}^{(i,d)}, \mathbf{\hat{Y}}^{(i,d)}\right) = \frac{1}{48N_{\text{test}}} \sum_{i=1}^{N_{\text{test}}}\sum_{j=1}^{48}\left(Y^{(i,d)}_j-\hat{Y}^{(i,d)}_j\right)^2$,}
\end{equation}
and the relative mean squared prediction error (RMSPE)
\begin{equation} \label{metric2}
\resizebox{1.0\hsize}{!}{$\text{RMSPE}\left(\mathbf{Y}^{(i,d)}, \mathbf{\hat{Y}}^{(i,d)}\right) = \frac{1}{N_{\text{test}}}\sum_{i=1}^{N_{\text{test}}} \frac{\sum_{j=1}^{48}\left(Y^{(i,d)}_j-\hat{Y}^{(i,d)}_j\right)^2}{\sum_{j=1}^{48}\left(Y^{(i,d)}_j\right)^2}.$}
\end{equation}

The implementations of the considered models are summarized as follows. Note that the time-wise Z-score standardization described in Section \ref{sec3} is utilized in all models. For the multivariate linear regression (`Multi LR'), the half-hourly temperature measurements in a week are treated as individual features (i.e., the dimension of input is 336), which are jointly mapped to half-hourly electricity demands throughout the week (i.e., the dimension of output is 336) through $336$ linear functions. In this non-linear formulation, the temporal information within each day is not efficiently captured. The number of unknown parameters is 112,896 (i.e., $48\times7\times48\times7$). For the LSTM-based sequential learning model (`Seq2Seq LSTM'), we use a LSTM layer with the input shape being $(48, 7)$ and the output is a $336$-dimensional vector. This layer is followed by a fully connected layer of $336$ neurons. The output is re-shaped into $(48, 7)$ to create the outputs. As for the functional models, first, the multivariate functional principal component analysis is numerically implemented for both the functional covariates and functional responses, based on the procedure in Section \ref{sec3.3}. We choose the number of functional principal components using the fraction of variance explained approach, with the cutoff being $99\%$. The selected values are $\hat{L}$=$11$ and $\hat{P}$=$10$. It means that, for any given subject, the complete information among the 7 correlated temporal covariates are well preserved in the $11$-dimensional vector that holds the projection of covariates onto the principal components. Analogously, the 7 correlated responses are well represented by the projections. The function-on-function linear model uses linear models to learn the unknown parameters in the $10\times 11$ transformation matrix. To build the proposed model in Eq \eqref{formulation_new2}, we build a neural network with the following architecture: 11 nodes in the input layer, 16 neurons in the first hidden layer with `elu' (i.e., Exponential Linear Unit) being the activation function, and 10 nodes in the output layer with linear activation functions. 

The results including the number of parameters and the accuracy metrics are given in Table \ref{tab1}. It can be seen that the proposed model outperforms the other models. Note that the observed time series are regular in this experiment. Therefore, the error of the non-functional models reported in the table does not include the possible biases introduced when transforming the raw data into regular time series.

\subsection{An Application to Short-term Traffic Predictions}\label{sec4.2}
Making short-term predictions for traffic is a substantial problem in the transportation field. Accurate predictions of traffic not only help the drivers make smarter choices that can save them time and fuel but also assist authorities in managing the transportation system. Characteristic metrics for the traffic situation at a given location often include speed (i.e., average vehicle travel distance per unit time), flow (i.e., number of vehicles passed per unit time), and occupancy (i.e., percentage of time in which a unit length of roadway is occupied by vehicles) \cite{chiou2016multivariate}. In this experiment, we use the proposed model to simultaneously predict the trajectories of the three traffic metrics in the remaining time of the day, based on the partially observed trajectories up to a certain time.  

We crawled data from the Caltrans Performance Measurement System (PeMS), a publicly accessible system that contains real-time traffic data from over 39,000 individual detectors spanned over California's highway system. Specifically, we obtain the historical speed, flow, and occupancy data from detector `400017', which is located on the southbound of CA-85 (near Bascom Avenue in Los Gatos). Totally, we include 739 non-holiday workdays between 1/1/2017 and 12/31/2019. All the three traffic variables were recorded over the day in 5-min intervals. The problem is to predict the speed, flow and occupancy over $\mathcal{T}=[10:00, 24:00]$ using the corresponding trajectories within $\mathcal{S}=[00:00, 10:00]$.

\begin{figure}[htbp]
	\begin{subfigure}[t]{3.25in}
		\centering
        \caption{One example in the testing set.}\label{temp}
		\includegraphics[width=8.75cm]{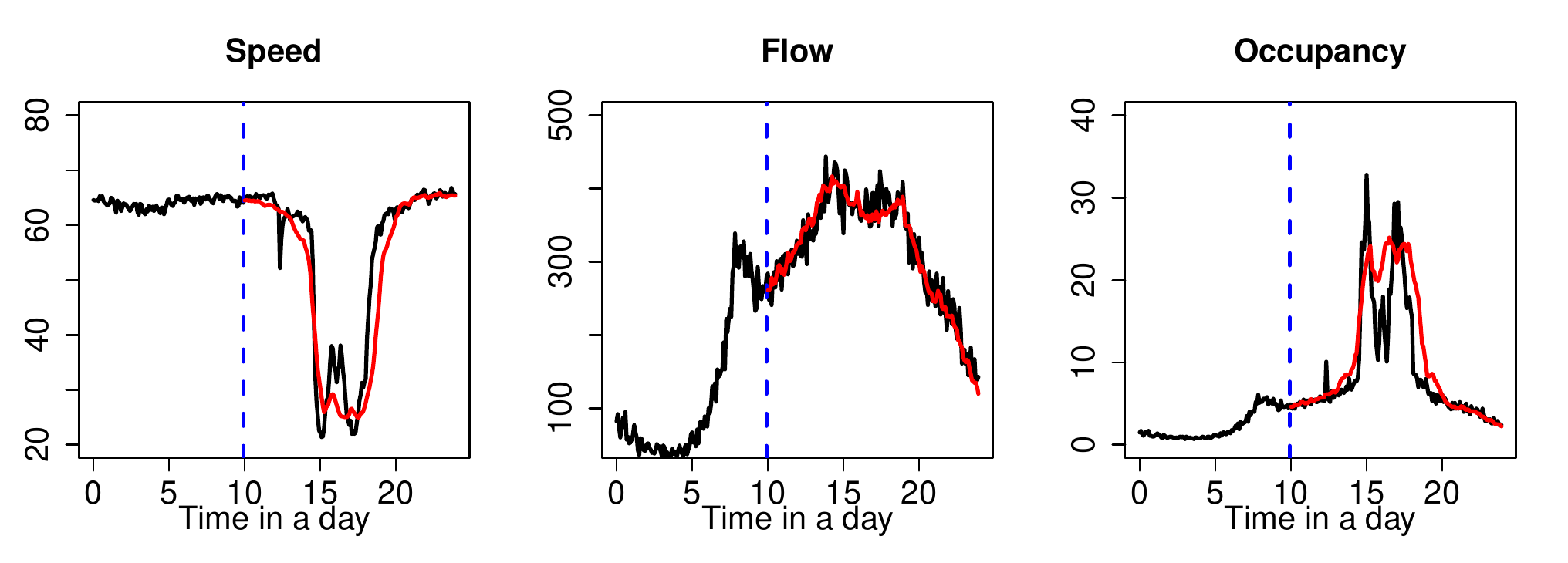}		
	\end{subfigure}
    \par\bigskip
	\begin{subfigure}[t]{3.25in}
		\centering
        \caption{Another example in the testing set.}\label{demand}
		\includegraphics[width=8.75cm]{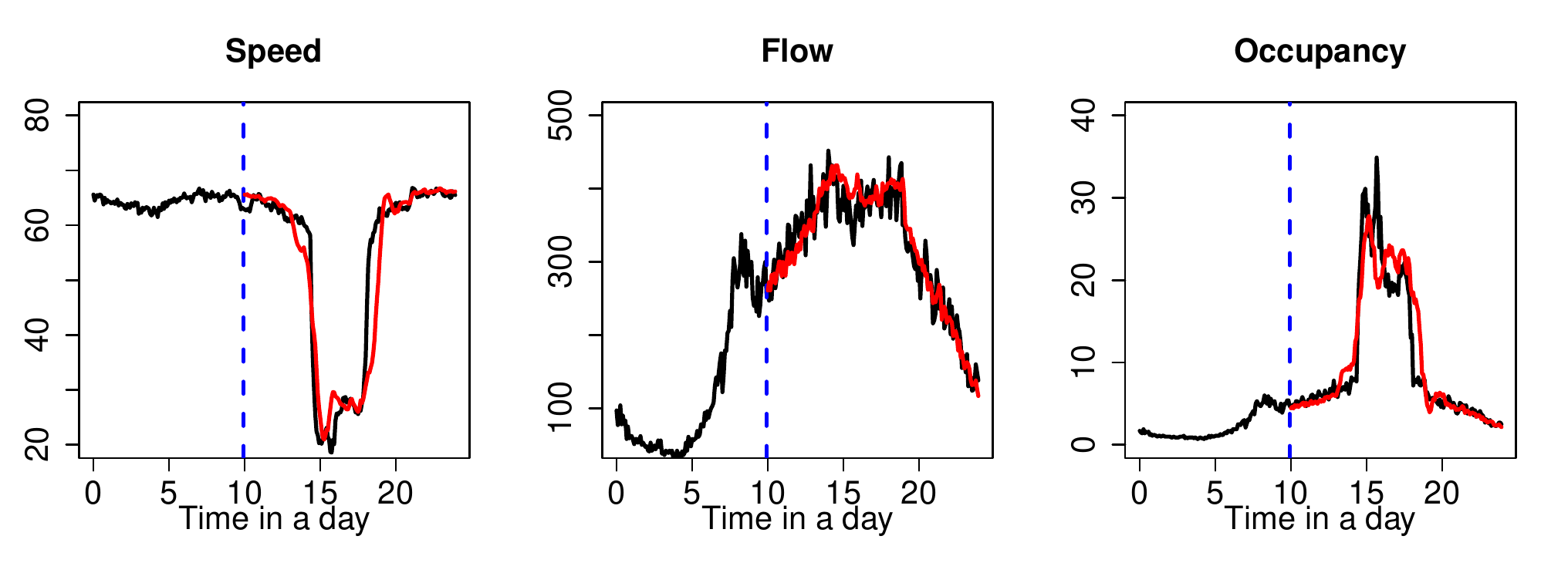}		
	\end{subfigure}
	\caption{Two illustrative examples from the testing set. Trajectories before the blue vertical line are the inputs. The black trajectories after the blue vertical line are the ground truth. The red trajectories are the predicted trajectories from the proposed mode. }\label{dataset2}
\end{figure}

We randomly assign $80\%$ of the samples into the training set and the remaining $20\%$ into the testing set. The implementation of the models are similar to Section \ref{sec4.1}. There are $120$ observations in period $\mathcal{T}$ and $168$ measurements over $\mathcal{S}$ for each traffic variable. So the number of unknown parameters for the multivariate linear regression models is $181,440$ (i.e., $120\times3\times168\times3$). The selected number of principal components are $\hat{L}$=$27$ and $\hat{P}$=$30$, based on the $99\%$ variance of explained rule. For the proposed model, there is one hidden layer with 16 neurons. As shown by Table \ref{tab2}, the proposed model yields the best performance in terms of both RMSE and RMSPE. The inputs, the actual responses, and the predictions from our model for two randomly picked subjects in the testing set are visualized in Fig. \ref{dataset2}. Compare to the result in  Section \ref{sec4.1}, the achieved improvement over the linear function-on-function model is smaller. Our explanation is that the underlying relationship is close to a linear mapping in this study, while the mapping in Section \ref{sec4.1} is complex.

\begin{table*}[htbp]
\caption{RMSE and RMSPE comparisons for the electricity analysis task in Section \ref{sec4.1}.}
\begin{center}
\begin{tabular}{ccccc}
\hline
\hline
\textbf{Model}& \textbf{$\#$ of parameters}&   \textbf{RMSE for each response} & \textbf{RMSPE for each response}\\
\hline
Multi LR& 112,896 & (442.9, 383.5, 339.5, 323.8, 372.7, 343.2, 363.2)
 & (0.29, 0.25, 0.22, 0.21, 0.25, 0.12, 0.13) \\
Seq2Seq LSTM & 575,568 & (176.2, 152.8, 153.2, 156.7, 168.6, 157.5, 158.6)
& (0.12, 0.10, 0.10, 0.11, 0.12, 0.11, 0.11) \\
FFLM & 110 & (210.9, 182.1, 184.5, 194.9, 198.3, 170.1. 171.0) &(0.14, 0.12, 0.12, 0.13, 0.14, 0.12, 0.13)

 \\
Proposed model & 362 & \textbf{(156.2, 133.7, 131.3, 139.7, 154.2, 122.1, 132.1)} &  \textbf{(0.11, 0.09, 0.09, 0.10, 0.11, 0.09, 0.10)}
\\
\hline
\hline

\end{tabular}
\label{tab1}
\end{center}

\bigskip

\caption{RMSE and RMSPE comparisons for the short-term traffic prediction task in Section \ref{sec4.2}.}
\begin{center}
\begin{tabular}{ccccc}
\hline
\hline
\textbf{Model}& \textbf{$\#$ of parameters}&   \textbf{RMSE for each response} & \textbf{RMSPE for each response}\\
\hline
Multi LR& 181,440 &(108.86, 10.65, 16.94)
 & (0.344, 0.714, 0.322) \\
Seq2Seq LSTM & 1,278,648 & (40.21, 4.62, 8.13)
& (0.154, 0.387, 0.169) \\
FFLM & 810 & (36.34, 4.42, 7.54)
 &(0.131, 0.337, 0.140)

 \\
Proposed model & 958 & \textbf{(34.62, 3.98, 6.63)
} &  \textbf{(0.125, 0.307, 0.122)}
\\
\hline
\hline
\end{tabular}
\label{tab2}
\end{center}
\end{table*}

\section{Conclusions}\label{sec5}
In this paper, we proposed a novel model for the multivariate time series regression problem, a frequently encountered topic in a wide range of fields. Compared to the existing approaches, the proposed model possesses several advantages, including its ability to handle versatile format of time series data, capture timely varying correlations among variables, and build complex mappings. To enhance the understanding, we described the numerical implementation of the proposed model step-by-step. We applied the proposed model to study the association between daily temperature and electricity demand in a week and to tackle short-term traffic prediction problem, in comparison with several common practices in the prior art. The proposed non-linear functional model produced smaller estimation and prediction errors than the state-of-the-art approaches. We expect the proposed model to be widely applicable to diverse real-world problems where the goal is to study the correlation among several time series.



\balance

\bibliographystyle{IEEEtran}
\bibliography{fanova}

\end{document}